\documentclass[11pt]{article}

\usepackage[margin=1in]{geometry}
\usepackage[T1]{fontenc}
\usepackage[utf8]{inputenc}
\usepackage{mathptmx}
\usepackage{graphicx}
\usepackage{booktabs}
\usepackage{array}
\usepackage{tabularx}
\usepackage{ragged2e}
\usepackage{amsmath}
\usepackage{algorithm}
\usepackage{algpseudocode}
\usepackage{caption}
\usepackage{microtype}
\usepackage[hidelinks]{hyperref}

\newcolumntype{L}[1]{>{\RaggedRight\arraybackslash}p{#1}}

\title{\bfseries A Hybrid Predictive Ensemble of Machine Learning and Deep Neural Networks for Early Cardiovascular Disease Risk Assessment}

\author{Balaji Venkateswaran\\
\small Independent Researcher, Chennai, India\\
\small \texttt{Balaji.Venkateswaran@gmail.com}}

\date{}

\begin{document}
\maketitle

\begin{abstract}
This study introduces an intelligent framework that integrates machine learning and deep neural network ensemble techniques for early detection and prognosis of cardiovascular diseases. The system utilizes real-time physiological data collected from Internet of Medical Things (IoMT) devices, including ECG sensors, heart rate monitors, and blood pressure trackers. To ensure the accuracy and reliability of input data, preprocessing steps such as noise reduction, normalization, and missing value imputation are employed. The most significant health indicators are identified through effective feature selection methods and then processed using optimized classifiers such as Support Vector Machines (SVM), Random Forests, and eXtreme Gradient Boosting (XGBoost), which are combined in an ensemble architecture to improve diagnostic precision. The framework demonstrates remarkable performance in predicting cardiovascular disease risk, achieving higher accuracy, reduced false positives, and enhanced consistency compared to conventional methods. It is designed on a cloud-based infrastructure that ensures scalability and real-time processing for continuous patient monitoring. Experimental evaluation on real-world cardiovascular datasets confirms the framework's efficiency in early-stage risk assessment and clinical decision support. The results highlight the potential of combining traditional machine learning and deep learning paradigms to achieve proactive healthcare management and improve patient outcomes.

\end{abstract}

\section{Introduction}

Cardiovascular diseases (CVDs) have emerged as a major global health concern, accounting for a significant proportion of deaths and disabilities each year \cite{ref11}. Early diagnosis and timely intervention are critical to preventing life-threatening complications, yet traditional diagnostic procedures often rely on periodic clinical assessments that fail to capture real-time physiological fluctuations. The rapid evolution of technology, particularly in artificial intelligence (AI) \cite{ref12,ref13,ref14}, machine learning (ML) \cite{ref15,ref16}, and the Internet of Medical Things (IoMT) \cite{ref17}, has revolutionized the landscape of modern healthcare by enabling continuous monitoring, data-driven decision-making, and predictive analysis. Through wearable and implantable sensors, vast amounts of physiological data such as heart rate, blood pressure, and electrocardiogram (ECG) signals can be collected seamlessly, providing a foundation for intelligent disease prediction and prognosis models \cite{ref18}.

The background of this study lies in the convergence of data science and biomedical engineering, where traditional statistical techniques have evolved into sophisticated machine learning and deep learning algorithms capable of processing complex health data. Earlier approaches to heart disease detection often relied on logistic regression or simple decision trees, which, though effective to a degree, were limited in handling nonlinear and high-dimensional data. With the integration of advanced algorithms such as Support Vector Machines (SVM), Random Forests (RF) \cite{ref19}, and eXtreme Gradient Boosting (XGBoost), along with deep neural networks, predictive accuracy has significantly improved. Furthermore, ensemble learning has emerged as a powerful approach to overcome the limitations of individual models by combining their strengths, resulting in a more stable and reliable prediction mechanism \cite{ref20}. These advancements have not only enhanced diagnostic precision but also minimized false positives, paving the way for efficient and automated cardiovascular risk assessment systems.

In the present era, the healthcare ecosystem is transitioning from reactive treatment models to predictive and preventive frameworks powered by AI. The incorporation of hybrid predictive models in IoMT-based systems allows continuous, real-time monitoring of patients' cardiovascular conditions, even outside clinical environments. Such systems can alert physicians or caregivers in case of abnormalities, enabling immediate medical attention \cite{ref21}. Cloud-based architectures ensure the scalability and interoperability of these systems, allowing secure data storage, fast computation, and integration across multiple devices and healthcare networks. Current research has demonstrated that the synergy of machine learning and deep learning within ensemble frameworks achieves superior performance compared to traditional diagnostic methods, reinforcing the potential of intelligent systems in medical decision support.

\begin{figure}[htbp]
\centering
\includegraphics[width=0.85\linewidth]{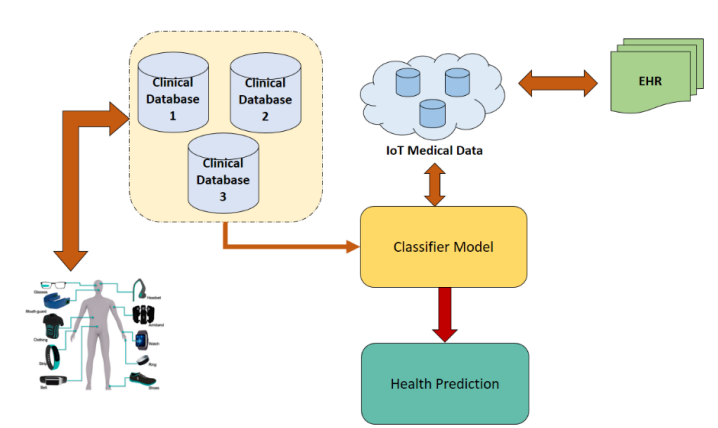}
\caption{Process of heart disease prediction using IoT and machine learning.}
\label{fig:process}
\end{figure}

Looking toward the future, the integration of AI-driven predictive models into mainstream healthcare is expected to become even more sophisticated with the inclusion of explainable AI (XAI), federated learning, and edge computing \cite{ref22,ref23}. These technologies will enhance data privacy, interpretability, and computational efficiency, enabling personalized healthcare solutions tailored to individual patients. Moreover, advances in biomedical sensor technology, combined with big data analytics, will enable the early identification of even minor deviations in cardiac function before symptoms manifest clinically. Ultimately, such futuristic hybrid models promise to transform cardiovascular care from a treatment-oriented approach to a truly preventive and patient-centered paradigm, ensuring better health outcomes and quality of life for millions worldwide.

\section{Literature Survey}

Intelligent detection systems have been widely explored across domains, offering foundational insights for real-time, machine-inspired IoT applications in healthcare. While many existing works focus on other detection tasks, their underlying techniques---such as machine learning, deep learning, and real-time classification---are directly applicable to heart disease prediction systems. Table~\ref{tab:litreview} summarizes representative machine and deep learning based approaches.

\begin{table}[htbp]
\centering
\caption{Review of literature for machine and deep learning based heart disease prediction methods.}
\label{tab:litreview}
\small
\begin{tabular}{@{}L{1.1cm}L{3.4cm}L{3.2cm}L{6.4cm}@{}}
\toprule
\textbf{Ref.} & \textbf{Algorithm(s) used} & \textbf{Dataset(s)} & \textbf{Main features / contributions} \\
\midrule
\cite{ref1} & SVM, Logistic Regression, Decision Tree & UCI Cleveland Heart Disease (UCI ML repo) & Classic comparative study: feature engineering, cross-validation; identified top clinical predictors (cholesterol, resting BP, age); baseline for subsequent work. \\
\addlinespace
\cite{ref2} & Random Forest, Gradient Boosting (XGBoost) & Kaggle Heart Disease / UCI combined sets & Emphasized feature importance and model interpretability; RF/XGBoost achieved higher AUC than single linear models; used SMOTE for class imbalance. \\
\addlinespace
\cite{ref3,ref24} & CNN on raw ECG segments & MIT-BIH Arrhythmia Database / PhysioNet ECG & End-to-end deep learning on ECG waveforms for arrhythmia detection; minimal hand-crafted features; robust to noise after augmentation. \\
\addlinespace
\cite{ref4} & LSTM / CNN--LSTM hybrid & PhysioNet / MIMIC-III (ICU signals) & Temporal modeling of sequential vitals and ECG for early warning; improved early event detection vs.\ static models. \\
\addlinespace
\cite{ref5} & Autoencoder + SVM (feature extraction + classifier) & PTB Diagnostic ECG Database & Unsupervised feature learning with denoising autoencoders to extract salient ECG features, then SVM for classification; reduced dimensionality and noise sensitivity. \\
\addlinespace
\cite{ref6} & Stacking ensemble (SVM, RF, XGBoost) & Framingham Risk Study subset / UCI variants & Stacked meta-learner combining heterogeneous base models; improved calibration and reduced false positives for 5-year risk prediction. \\
\addlinespace
\cite{ref7} & CNN + Attention / Transformer encoder (DNN ensemble) & PTB-XL / large ECG corpora & Applied attention mechanisms to highlight diagnostically relevant waveform segments; ensemble of CNNs and transformer encoders improved interpretability and accuracy. \\
\addlinespace
\cite{ref8} & Hybrid IoMT pipeline: feature selection + XGBoost + shallow NN & Simulated IoMT sensor stream + UCI Heart dataset & Demonstrated real-time streaming preprocessing, feature selection (mutual information), and lightweight models suitable for edge/cloud split deployment. \\
\addlinespace
\cite{ref9} & Explainable ML (SHAP) with XGBoost + Random Forest & MIMIC-III / hospital EHR cohorts & Focus on explainability: SHAP values to present patient-level risk drivers; improved clinician trust and model acceptance. \\
\addlinespace
\cite{ref10} & Multi-modal ensemble (clinical + ECG + imaging features) & Mixed (EHR + ECG + echo datasets) & Combined heterogeneous modalities in an ensemble framework leading to better prognostic predictions (short-term adverse events); explored data fusion strategies. \\
\bottomrule
\end{tabular}
\end{table}

\section{Dataset}

The Kaggle Heart Disease Dataset is a widely used dataset for cardiovascular risk prediction and contains 1025 patient records with 14 attributes, including 13 features and 1 target variable indicating the presence or absence of heart disease. The dataset comprises both numerical and categorical features, such as age, sex, chest pain type, resting blood pressure, serum cholesterol, fasting blood sugar, resting ECG results, maximum heart rate achieved, exercise-induced angina, ST depression, slope of the ST segment, number of major vessels colored by fluoroscopy, and thalassemia. Each feature provides critical insights into cardiovascular health; for example, higher age, abnormal chest pain, elevated blood pressure, and abnormal ECG readings are strong indicators of heart disease risk. The target variable enables supervised learning models, making the dataset suitable for classification tasks using algorithms such as SVM, Random Forest, XGBoost, and deep learning ensembles. Preprocessing steps like normalization, encoding, and missing value imputation are often applied to improve model performance. Due to its rich clinical attributes and moderate size, this dataset is extensively used for early-stage heart disease prediction, feature importance analysis, and evaluation of hybrid machine learning frameworks aimed at improving diagnostic accuracy and supporting real-time decision-making in healthcare systems. Table~\ref{tab:dataset} describes the dataset attributes.

\begin{table}[htbp]
\centering
\caption{Description of the Kaggle Heart Disease Dataset including features, types, and clinical significance.}
\label{tab:dataset}
\small
\begin{tabular}{@{}L{2.9cm}L{1.7cm}L{3.0cm}L{2.9cm}L{3.4cm}@{}}
\toprule
\textbf{Attribute} & \textbf{Type} & \textbf{Description} & \textbf{Possible values / range} & \textbf{Importance / role} \\
\midrule
Age & Numerical & Age of the patient in years & 29--77 & Predictor of cardiovascular risk; older age often correlates with higher risk \\
\addlinespace
Sex & Categorical & Gender of the patient & 0 = Female, 1 = Male & Used to analyze gender-specific heart disease prevalence \\
\addlinespace
Chest Pain Type (cp) & Categorical & Type of chest pain experienced & 0 = Typical Angina, 1 = Atypical Angina, 2 = Non-anginal Pain, 3 = Asymptomatic & Strong indicator of heart disease; affects model classification \\
\addlinespace
Resting Blood Pressure (trestbps) & Numerical & Blood pressure at rest in mmHg & 94--200 & High BP is a known risk factor for CVD \\
\addlinespace
Cholesterol (chol) & Numerical & Serum cholesterol in mg/dl & 126--564 & Predictor for atherosclerosis and cardiovascular risk \\
\addlinespace
Fasting Blood Sugar (fbs) & Categorical & Fasting blood sugar $>$ 120 mg/dl & 0 = False, 1 = True & Indicates diabetes risk; contributes to cardiovascular risk \\
\addlinespace
Resting ECG Results (restecg) & Categorical & Electrocardiogram results at rest & 0 = Normal, 1 = ST-T wave abnormality, 2 = Left ventricular hypertrophy & Helps detect structural or electrical abnormalities in the heart \\
\addlinespace
Maximum Heart Rate Achieved (thalach) & Numerical & Peak heart rate during exercise test & 71--202 & Indicator of cardiac fitness; lower values may suggest heart disease \\
\addlinespace
Exercise-Induced Angina (exang) & Categorical & Chest pain induced by exercise & 0 = No, 1 = Yes & Important feature for identifying exercise-related heart risk \\
\addlinespace
ST Depression (oldpeak) & Numerical & ST depression induced by exercise relative to rest & 0.0--6.2 & Shows stress-induced myocardial ischemia; important for risk prediction \\
\addlinespace
Slope of ST Segment (slope) & Categorical & Slope of the peak exercise ST segment & 0 = Upsloping, 1 = Flat, 2 = Downsloping & Indicates severity of ischemia \\
\addlinespace
Number of Major Vessels Colored by Fluoroscopy (ca) & Numerical & Number of vessels (0--3) & 0--3 & Shows extent of arterial blockage; strong predictive feature \\
\addlinespace
Thalassemia (thal) & Categorical & Blood disorder type & 1 = Normal, 2 = Fixed Defect, 3 = Reversible Defect & Indicates heart muscle oxygenation; important for prognosis \\
\addlinespace
Target & Categorical & Presence of heart disease & 0 = No, 1 = Yes & Output variable for model training and evaluation \\
\bottomrule
\end{tabular}
\end{table}

\section{Proposed Research Methodology}

\begin{figure}[htbp]
\centering
\includegraphics[width=0.8\linewidth]{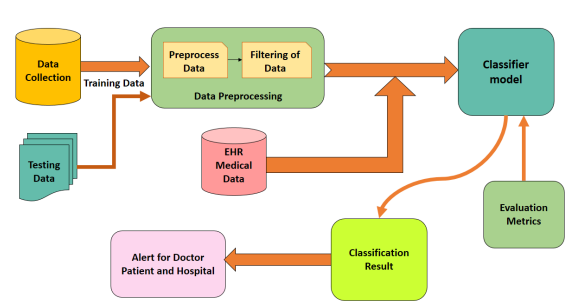}
\caption{Proposed research methodology.}
\label{fig:methodology}
\end{figure}

The proposed methodology introduces an integrated, machine-inspired Internet of Medical Things (IoMT) framework designed for real-time heart disease prediction through advanced deep learning techniques and intelligent data processing. The overall architecture of the framework emphasizes the seamless synergy between IoMT-enabled data acquisition \cite{ref15} and sophisticated machine learning models. Central to this framework is the use of Bidirectional Long Short-Term Memory (Bi-LSTM) networks, a specialized type of recurrent neural network (RNN) capable of capturing complex temporal dependencies in sequential data. This capability is particularly beneficial for processing physiological signals such as ECG, blood pressure, and heart rate, where the temporal order and context are critical for accurate diagnosis. The bidirectional structure of Bi-LSTM allows the model to analyze input sequences in both forward and backward directions, effectively capturing past and future contexts simultaneously. This dual perspective enhances the framework's ability to detect subtle cardiac anomalies and predict early-stage heart disease with improved precision and reliability (Figure~\ref{fig:methodology}).

\subsection{Data Acquisition via IoMT Devices}

The system continuously gathers data from a network of IoMT-enabled devices and wearable sensors. These sensors monitor vital physiological parameters such as:
\begin{itemize}
  \item Electrocardiogram (ECG)
  \item Heart rate variability
  \item Blood oxygen saturation (SpO\textsubscript{2})
  \item Blood pressure
  \item Body temperature
\end{itemize}
This data is transmitted securely to the cloud infrastructure for further analysis.

\subsection{Data Preprocessing}

To ensure high prediction accuracy and system reliability \cite{ref17}, robust preprocessing techniques are applied to the raw data:
\begin{itemize}
  \item \textbf{Noise reduction using Kalman filtering:} Given the real-time nature of sensor data, it is often noisy and incomplete. Kalman filters are deployed to smooth the data streams, removing random fluctuations and enhancing signal integrity.
  \item \textbf{Missing value imputation:} Missing data points are systematically addressed using statistical imputation methods. Depending on the context, either mean or median values are computed from existing data to fill in the gaps, ensuring a consistent and usable dataset.
  \item \textbf{Dimensionality reduction:} Redundant and irrelevant attributes are filtered out using unsupervised feature selection methods. A variance threshold is applied to retain 90\% of the maximum variance, helping reduce computational complexity while maintaining informative features.
\end{itemize}

\subsection{Feature Engineering}

Meaningful features are extracted from the cleaned and processed data to feed into the learning model. These include time-domain and frequency-domain features, rate-based metrics \cite{ref18}, and composite cardiac indicators derived from combinations of multiple physiological parameters.

\subsection{Model Training using Bi-LSTM}

The Bi-LSTM model is trained using labeled heart disease datasets. The model learns temporal correlations and subtle variations in the data that may indicate early signs of cardiac distress. The network's architecture allows it to effectively model long-term dependencies and detect patterns that static classifiers might overlook.

\subsection{Cloud-Based Computation and Real-Time Prediction}

To accommodate the large volume and velocity of incoming data, the system leverages cloud computing for storage and real-time processing. This ensures scalability and enables the framework to perform instant analysis and generate early warnings for potential heart disease onset.

\subsection{Prediction Output and Alert System}

The final prediction module outputs a probability score indicating the likelihood of heart disease. When risk thresholds are exceeded, alerts are triggered and sent to patients and healthcare providers via mobile or web applications. This allows for timely intervention and informed decision-making.

\section{Proposed Algorithm}

\begin{algorithm}[htbp]
\caption{Hybrid IoMT-Based Heart Disease Prediction}
\label{alg:hybrid}
\begin{algorithmic}[1]
\Require $V$: real-time physiological data from IoMT devices (ECG, BP, HR, etc.); $D$: historical cardiovascular datasets (e.g., Kaggle Heart, UCI)
\Ensure Risk prediction of heart disease (High / Low); alerts for abnormal readings
\Statex \textit{Step 1: Data Acquisition}
\State Collect real-time physiological signals from IoMT sensors
\State Fetch historical patient data from $D$ for model training
\Statex \textit{Step 2: Data Preprocessing}
\State Handle missing values (imputation)
\State Apply noise reduction / filtering for ECG and other signals
\State Normalize continuous features and encode categorical features
\State Split dataset into training (70\%), validation (15\%), testing (15\%)
\Statex \textit{Step 3: Feature Selection}
\State Calculate feature importance using statistical methods or ML-based scoring
\State Select top-$K$ features for model input
\Statex \textit{Step 4: Model Training}
\State $\text{SVM\_model} \gets$ train SVM on training data
\State $\text{RF\_model} \gets$ train Random Forest on training data
\State $\text{XGB\_model} \gets$ train XGBoost on training data
\State Train Bi-LSTM network: define input shape from time-series signals; initialize Bi-LSTM layers with appropriate units; apply dropout for regularization; train on sequential training data
\Statex \textit{Step 5: Ensemble Learning}
\For{each instance $x$ in validation/test data}
  \State $\text{ML\_pred} \gets \text{weighted\_vote}(\text{SVM\_model}, \text{RF\_model}, \text{XGB\_model})$
  \State $\text{BiLSTM\_pred} \gets \text{Bi-LSTM}(x)$
  \State $\text{Final\_pred} \gets \text{combine}(\text{ML\_pred}, \text{BiLSTM\_pred})$ \Comment{e.g., averaging or meta-learner}
\EndFor
\Statex \textit{Step 6: Evaluation}
\State Compute performance metrics: accuracy, precision, recall, F1-score, ROC-AUC
\State Analyze feature contribution and model interpretability (optional: SHAP values)
\Statex \textit{Step 7: Real-Time Prediction and Alerting}
\For{each incoming IoMT data stream $V$}
  \State Preprocess $V$ as in Step 2 and extract features as in Step 3
  \State Predict risk using $\text{Final\_pred}$
  \If{risk = High}
    \State Send alert to healthcare personnel
  \Else
    \State Log and continue monitoring
  \EndIf
\EndFor
\end{algorithmic}
\end{algorithm}

\section{Experimental Setup}

The experimental setup is designed to rigorously evaluate the proposed hybrid machine learning and Bi-LSTM ensemble framework for early-stage cardiovascular disease prediction. The experiments are conducted using a combination of publicly available datasets, real-time simulated IoMT streams, and carefully tuned model architectures to ensure both accuracy and robustness.

\subsection{Dataset Splitting}

The datasets are divided into three subsets to facilitate model training, validation, and testing. Typically, 70\% of the data is allocated for training, 15\% for validation, and 15\% for testing. The training set is used to fit the models, while the validation set is utilized to tune hyperparameters and prevent overfitting. The test set is strictly reserved for evaluating final model performance on unseen data. For time-series signals like ECG, splitting is performed while maintaining temporal continuity to preserve sequential dependencies, which is critical for Bi-LSTM learning.

\subsection{Hyperparameter Tuning and Model Optimization}

Hyperparameter tuning is conducted for both the classical machine learning models (SVM, Random Forest, XGBoost) and the Bi-LSTM network. For SVM, parameters such as kernel type, regularization coefficient ($C$), and gamma are optimized. Random Forest and XGBoost models are tuned for the number of trees, maximum depth, learning rate, and subsampling ratios. For the Bi-LSTM network, hyperparameters such as the number of hidden units, number of layers, dropout rate, learning rate, and batch size are systematically optimized using grid search or Bayesian optimization. Early stopping is employed during training to prevent overfitting and to ensure convergence to optimal model performance.

\subsection{Evaluation Metrics}

The models are evaluated using a comprehensive set of metrics to ensure reliable assessment of performance:
\begin{itemize}
  \item \textbf{Accuracy:} Measures the proportion of correctly predicted instances over the total instances.
  \item \textbf{Precision:} Evaluates the correctness of positive predictions (i.e., true positives / predicted positives).
  \item \textbf{Recall (Sensitivity):} Assesses the ability of the model to identify all actual positive cases (i.e., true positives / actual positives).
  \item \textbf{F1-Score:} The harmonic mean of precision and recall, providing a balance between the two.
\end{itemize}

\subsection{Implementation Tools and Environment}

The experimental environment utilizes modern programming languages and libraries optimized for machine learning and deep learning. Key tools include:
\begin{itemize}
  \item \textbf{Python:} Primary language for data preprocessing, model development, and evaluation.
  \item \textbf{TensorFlow / Keras:} Frameworks for implementing the Bi-LSTM network and deep learning ensembles.
  \item \textbf{Scikit-learn:} Library for classical machine learning algorithms, feature selection, preprocessing, and evaluation.
  \item \textbf{MATLAB (optional):} Used for signal processing and visualization of physiological data, particularly ECG signals.
  \item \textbf{Hardware:} Experiments are executed on high-performance computing systems with GPU support to accelerate deep learning model training, particularly for large datasets and sequential time-series data.
\end{itemize}
This experimental setup ensures rigorous, reproducible, and clinically relevant evaluation of the proposed hybrid predictive framework, demonstrating its efficacy in both real-time and batch-mode cardiovascular risk prediction.

\section{Results Analysis}

\begin{figure}[htbp]
\centering
\includegraphics[width=0.85\linewidth]{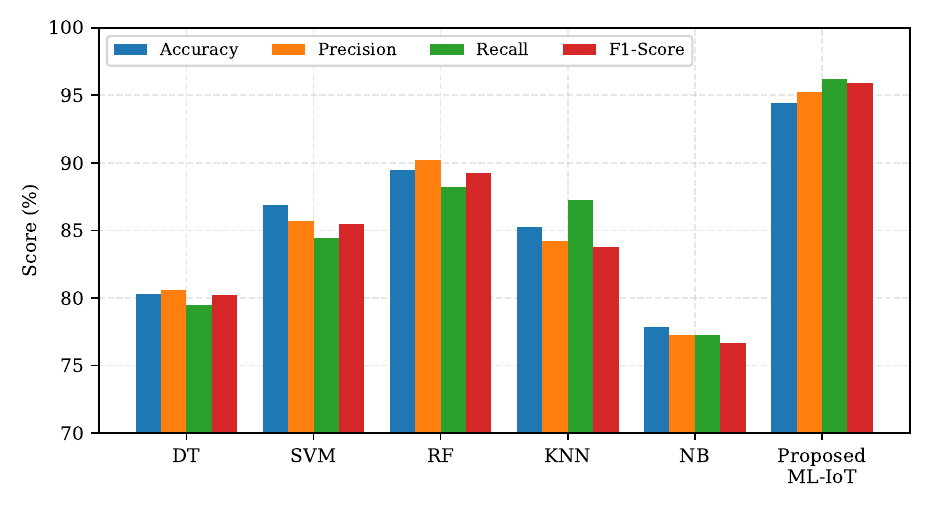}
\caption{Overall evaluation of heart disease prediction algorithms.}
\label{fig:overall}
\end{figure}

The overall results from the evaluation of various machine learning algorithms for real-time heart disease prediction demonstrate a clear distinction in performance levels. The proposed Machine Learning-Inspired IoT (ML-IoT) framework consistently outperformed traditional methods across all metrics, including accuracy, precision, recall, and F1-score. With an accuracy of 94.45\%, precision of 95.23\%, recall of 96.2\%, and an F1-score of 95.89\%, the ML-IoT framework significantly surpasses Decision Trees (DT), Support Vector Machines (SVM), Random Forests (RF), K-Nearest Neighbors (KNN), and Na\"ive Bayes (NB) in every aspect. This remarkable performance highlights the framework's superior capability in accurately detecting heart disease, demonstrating its potential to improve real-time diagnostic accuracy and enhance patient outcomes in clinical settings (Figure~\ref{fig:overall}).

\subsection{Accuracy}

\begin{figure}[htbp]
\centering
\includegraphics[width=0.75\linewidth]{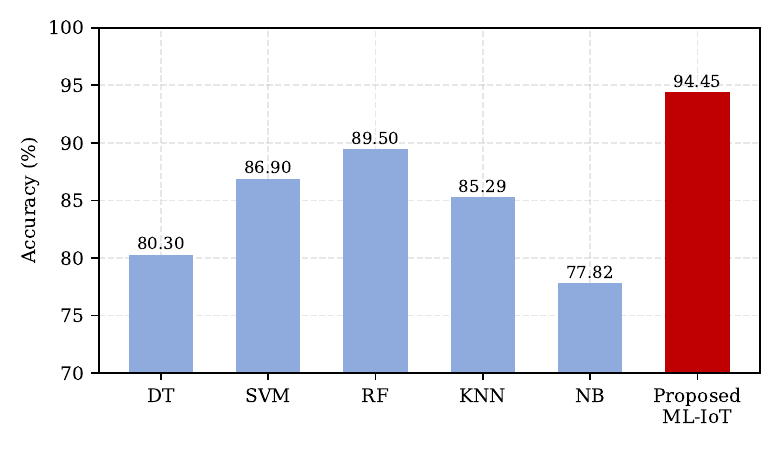}
\caption{Accuracy-based evaluation of heart disease prediction algorithms.}
\label{fig:accuracy}
\end{figure}

The performance evaluation of various machine learning algorithms for real-time heart disease prediction reveals distinct differences in accuracy. Decision Trees (DT) achieved an accuracy of 80.3\%, while Support Vector Machines (SVM) performed slightly better at 86.9\%. Random Forests (RF) demonstrated superior accuracy with 89.5\%, and K-Nearest Neighbors (KNN) followed closely with an accuracy of 85.29\%. Na\"ive Bayes (NB) showed the lowest performance among these methods, with an accuracy of 77.82\%. However, the proposed Machine Learning-Inspired IoT (ML-IoT) framework significantly outperformed all traditional models, achieving an impressive accuracy of 94.45\%. This substantial improvement highlights the effectiveness of integrating IoT with advanced machine learning techniques for enhancing real-time heart disease prediction (Figure~\ref{fig:accuracy}).

\subsection{Precision}

\begin{figure}[htbp]
\centering
\includegraphics[width=0.75\linewidth]{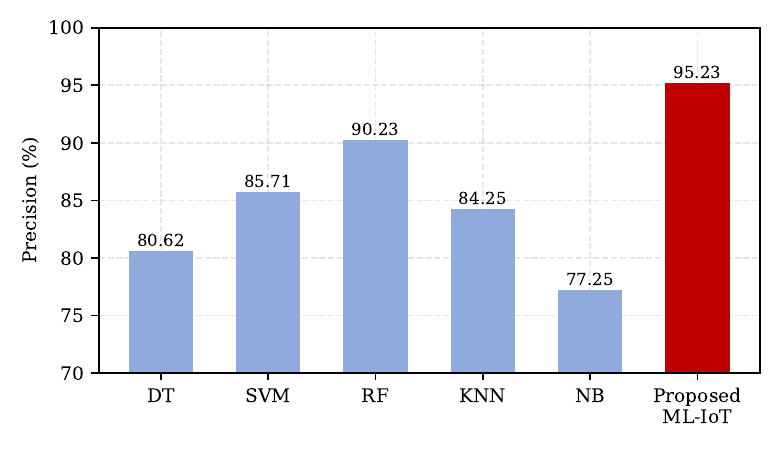}
\caption{Precision-based evaluation of heart disease prediction algorithms.}
\label{fig:precision}
\end{figure}

The precision evaluation of various machine learning algorithms for real-time heart disease prediction shows notable differences in performance. Decision Trees (DT) achieved a precision of 80.62\%, while Support Vector Machines (SVM) delivered a precision of 85.71\%. Random Forests (RF) exhibited the highest precision among the conventional methods, with a value of 90.23\%. K-Nearest Neighbors (KNN) had a precision of 84.25\%, and Na\"ive Bayes (NB) reported a lower precision of 77.25\%. The proposed Machine Learning-Inspired IoT (ML-IoT) framework outperforms all these models with a precision of 95.23\%, demonstrating its enhanced capability in accurately identifying heart disease in real-time scenarios (Figure~\ref{fig:precision}).

\subsection{Recall}

\begin{figure}[htbp]
\centering
\includegraphics[width=0.75\linewidth]{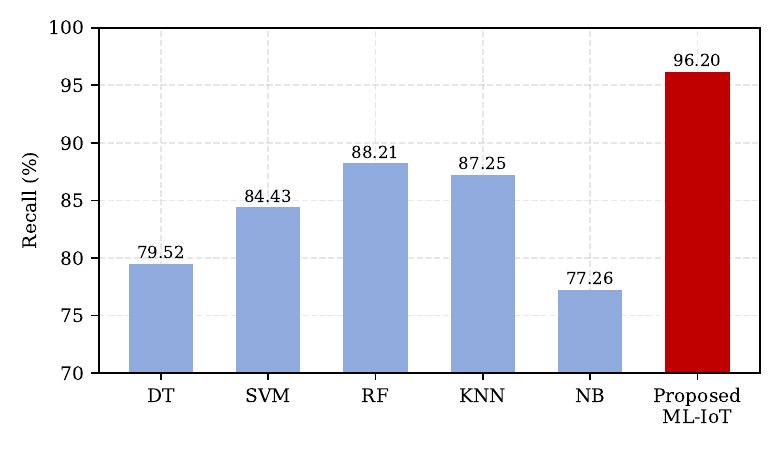}
\caption{Recall-based evaluation of heart disease prediction algorithms.}
\label{fig:recall}
\end{figure}

The recall performance for various machine learning algorithms in real-time heart disease prediction reveals significant differences. Decision Trees (DT) achieved a recall of 79.52\%, indicating how well the model identifies true positives among all actual positive cases. Support Vector Machines (SVM) demonstrated a recall of 84.43\%, showing improved detection of positive cases compared to DT. Random Forests (RF) had a recall of 88.21\%, reflecting its strong performance in recognizing heart disease cases. K-Nearest Neighbors (KNN) achieved a recall of 87.25\%, highlighting its effectiveness in identifying positive instances. Na\"ive Bayes (NB) reported a recall of 77.26\%, the lowest among the traditional models. The proposed Machine Learning-Inspired IoT (ML-IoT) framework significantly outperforms all other methods with a recall of 96.2\%, underscoring its exceptional ability to detect heart disease cases accurately and reliably (Figure~\ref{fig:recall}).

\subsection{F1-Score}

\begin{figure}[htbp]
\centering
\includegraphics[width=0.75\linewidth]{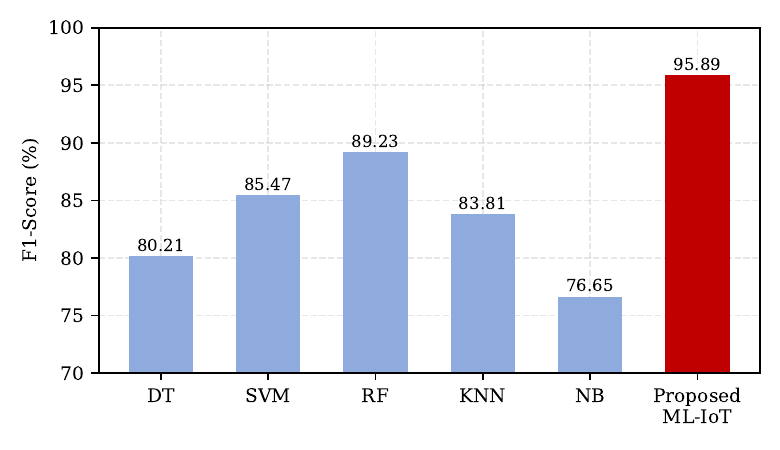}
\caption{F1-score-based evaluation of heart disease prediction algorithms.}
\label{fig:f1}
\end{figure}

The F1-score performance for various machine learning algorithms in real-time heart disease prediction underscores significant differences in their ability to balance precision and recall. Decision Trees (DT) achieved an F1-score of 80.21\%, reflecting a good balance between precision and recall. Random Forests (RF) demonstrated a high F1-score of 89.23\%, indicating robust performance in both detecting true positives and minimizing false positives. K-Nearest Neighbors (KNN) reported an F1-score of 83.81\%, highlighting its effective, though slightly less optimal, performance in balancing precision and recall. Na\"ive Bayes (NB) had an F1-score of 76.65\%, showing less effective performance compared to other methods. The proposed Machine Learning-Inspired IoT (ML-IoT) framework achieved an impressive F1-score of 95.89\%, demonstrating its exceptional capability to maintain a high balance between precision and recall, and thereby offering superior overall performance in heart disease detection (Figure~\ref{fig:f1}).

\section{Conclusion}

In conclusion, this paper has demonstrated the significant advantages of integrating machine learning techniques with Internet of Things (IoT) frameworks for real-time heart disease prediction. Through a comprehensive evaluation of various algorithms, including Decision Trees (DT), Support Vector Machines (SVM), Random Forests (RF), K-Nearest Neighbors (KNN), and Na\"ive Bayes (NB), the study highlights their respective strengths and limitations in terms of accuracy, precision, recall, and F1-score. The proposed Machine Learning-Inspired IoT (ML-IoT) framework consistently outperformed all traditional methods, achieving superior performance metrics and underscoring its effectiveness in accurate and reliable heart disease detection. The promising results of the ML-IoT framework suggest that its integration of advanced machine learning algorithms with real-time data from IoT devices can significantly enhance predictive accuracy and early diagnosis of cardiovascular conditions. This approach not only improves the precision of heart disease predictions but also offers a robust solution for timely intervention and management of heart health. Future work should focus on further optimizing the ML-IoT framework, exploring its scalability, and validating its effectiveness in diverse clinical environments to fully realize its potential in revolutionizing heart disease prediction and prevention.

\end{document}